\documentclass[twocolumn]{svjour3}          
\smartqed  
\usepackage{graphicx}
\usepackage{natbib}
\usepackage{amssymb}
\usepackage[font=small,labelfont=bf,labelsep=space]{caption}
\usepackage{floatrow}
\usepackage{tabularx}
\usepackage{color}
\usepackage{balance}

\usepackage{tabu}
\usepackage{xcolor}
\usepackage[cmex10]{amsmath}

\newcommand{\etal}{\textit{et~al}.}
\begin{document}

\title{Learning Human Pose Models from Synthesized Data for Robust RGB-D Action Recognition
}


\author{Jian Liu         \and
        Hossein Rahmani	 \and
        Naveed Akhtar \and
        Ajmal Mian 
}


\institute{J. Liu \at
			  \email{jian.liu@research.uwa.edu.au}
           \and
           H. Rahmani \at
              \email{hossein.rahmani@uwa.edu.au}
           \and
           N. Akhtar \at
           \email{naveed.akhtar@uwa.edu.au} 
           \and
           A. Mian \at
              \email{ajmal.mian@uwa.edu.au}
           \and
              CSSE, The University of Western Australia, 35 Stirling Highway, Crawley, WA 6009 \\
              Tel.: +61-8-64882702\\
              Fax: +61-8-64881089
}

\date{Received: date / Accepted: date}

\maketitle

\begin{abstract}
We propose Human Pose Models that represent RGB and depth images of human poses independent of clothing textures, backgrounds, lighting conditions, body shapes and camera viewpoints. Learning such universal models requires training images where all factors are varied for every human pose. Capturing such data is prohibitively expensive. Therefore, we develop a framework for synthesizing the training data. First, we learn representative human poses from a large corpus of real motion captured human skeleton data. Next, we fit synthetic 3D humans with different body shapes to each pose and render each from 180 camera viewpoints while randomly varying the clothing textures, background and lighting. Generative Adversarial Networks are employed to minimize the gap between synthetic and real image distributions. CNN models are then learned that transfer human poses to a shared high-level invariant space. The learned CNN models are then used as invariant feature extractors from real RGB and depth frames of human action videos and the temporal variations are modelled by Fourier Temporal Pyramid. Finally, linear SVM is used for classification. Experiments on three benchmark cross-view human action datasets show that our algorithm outperforms existing methods by significant margins for RGB only and RGB-D action recognition.
\keywords{Human Action Recognition \and Cross-view \and Cross-subject \and Depth Sensor \and CNN \and GAN}
\end{abstract}

\vspace{-7mm}
\section{Introduction}
\label{intro}
Human action recognition has many applications in security, surveillance, sports analysis, human computer interaction and video retrieval. However, automatic human action recognition algorithms are still challenged by noises due to action irrelevant factors such as changing camera viewpoints, clothing textures, body shapes, backgrounds and illumination conditions. In this paper, we address these challenges to perform robust human action recognition in conventional RGB videos and RGB-D videos obtained from range sensors.

A human action can be defined as a collection of sequentially organized human poses where the action is encoded in the way the human pose transitions from one pose to the other. However, for action classification, a human pose must be represented in a way that is invariant to the above conditions. Since some human poses are common between multiple actions and the space of possible human poses is much smaller compared to that of possible human actions, we first model the human pose independently and then model the actions as the temporal variations between human poses.

\begin{figure*}
\centering
  \includegraphics[width=1.0\textwidth]{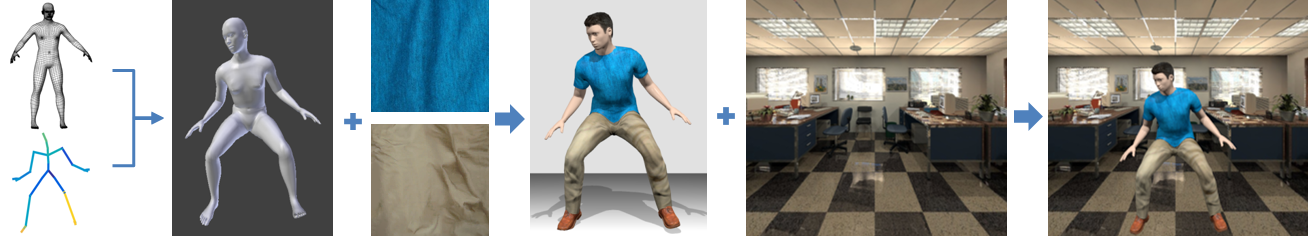}
\caption{Block diagram of the proposed synthetic RGB data generation. Representative human poses are learned from CMU MoCap skeleton database and a 3D human model is fitted to each skeleton. Four different 3D human body shapes are used. Clothing textures are randomly selected from a choice of 262 textures for shirts and 183 for trousers. Each model is placed in a random background, illuminated with three random intensity lamps and rendered from 180 camera viewpoints to generate RGB training images with known pose labels}
\label{fig:synthetic_gen}
\end{figure*}

To suppress action irrelevant information in videos, many techniques use dense trajectories \citep{nCTE, wang2011action, wang2013dense, wang2013action} or Hanklets \citep{Hankelets} which encode only the temporal cues that are essential for action classification. Such methods have shown good performance for human action recognition in conventional videos. However, they are still sensitive to viewpoint variations and do not fully exploit the appearance (human pose) information. Dense trajectories are also noisy and contain self occlusion artefacts.

While appearance is an important cue for action recognition, human poses appear very differently from different camera viewpoints. Research efforts have been made to model these variations. For example, synthetic 2D human poses from many viewpoints and their transitions were used for action recognition in \citep{lv2007single}. However, 3D viewpoint variations cannot be modelled accurately using 2D human poses. Spatio-temporal 3D occupancy grids built from multiple viewpoints were used in \citep{weinland2007action} to achieve view-invariant action recognition. However, occupancy grids rely on silhouettes which are noisy in real videos. In this paper, we use full 3D human models to learn a representation of the human pose that is not only invariant to viewpoint but also to other action irrelevant factors such as background, clothing and illumination.



Our contributions can be summarized as follows. Firstly, we propose a method for generating RGB and Depth images of human poses using Computer Graphics and Generative Adversarial Network (GAN) training. We learn representative human poses by clustering real human joint/skeleton data  obtained with motion capture technology (CMU MoCap database\footnote{http://mocap.cs.cmu.edu}). Each representative pose skeleton is fitted with synthetic 3D human models and then placed in random scenes, given different clothes, illuminated from multiple directions and rendered from 180 camera viewpoints to generate RGB and depth images of the human poses with known labels. Figure~\ref{fig:synthetic_gen} illustrates the proposed RGB data generation pipeline. Depth images are generated in a similar way except that they are devoid of texture and background. We employ GANs to minimize the gap between the distributions of  synthetic and real images. Although used as an essential component of network training in this work, the proposed synthetic data generation technique is generic and can be used to produce large amount of synthetic human poses for deep learning in general.

Secondly, we propose Human Pose Models (HPM) that are Convolutional Neural Networks and transfer human poses to a shared high level invariant space. The HPMs are trained with the images that are refined with GAN and learn to map input (RGB or Depth) images to one of the representative human poses irrespective of the camera viewpoint clothing, human body size, background and lighting conditions. The layers prior to the Softmax label in the CNNs serve as high-level invariant human pose representations. Lastly, we propose to temporally model the invariant human pose features with the Fourier Temporal Pyramid and use SVM for classification. The proposed methods work together to achieve robust RGB-D human action recognition under the modeled variations.

Experiments on three benchmark cross-view human action datasets show that our method outperforms existing state-of-the-art for action recognition in conventional RGB videos as well as RGB-D videos obtained from depth sensors. The proposed method improves RGB-D human action recognition accuracies by 15.4\%, 11.9\% and 8.4\% on the UWA3D-II \citep{HOPC2016PAMI}, NUCLA \citep{AOG} and NTU \citep{shahroudy2016ntu} datasets respectively. Our method also improves RGB human action recognition accuracy by 9\%.

This work is an extension of \citep{HPM+TM} where only depth image based human pose model and action recognition results were presented. However, depth images are almost always accompanied with RGB images. Therefore, we present the following extensions: (1) We present a method\footnote{The code for this method will be made public.} for synthesizing realistic RGB human pose images containing variations in clothing textures, background, lighting conditions and more variations in body shapes. This method can also synthesize depth images more efficiently compared to \citep{HPM+TM}. (2) We adopt Generative Adversarial Networks (GANs) to refine the synthetic RGB images and depth images so as to reduce their distribution gaps from real images and achieve improved accuracy. (3) We present a Human Pose Model HPM$_{\mathrm{RGB}}$ for human action recognition in conventional RGB videos which has wider applications. The proposed HPM$_{\mathrm{RGB}}$ achieves state-of-the-art human action recognition accuracy in conventional RGB videos. (4) We re-train the depth model HPM$_{\mathrm{3D}}$ using GoogleNet \citep{GoogLeNet} architecture which performs similar to the AlexNet \citep{AlexNet} model in \citep{HPM+TM} but with four times smaller feature dimensionality. (5) We perform additional experiments on the largest RGB-D human action dataset (NTU \citep{shahroudy2016ntu}) and report state-of-the-art results for action recognition in RGB videos and RGB-D videos in the cross-view and cross-subject settings. From here on, we refer to the Human Pose Models as HPM$_{\mathrm{RGB}}$ and HPM$_{\mathrm{3D}}$ for RGB and depth modalities respectively.

\section{Ralated Work}
The closest work to our method is the key pose matching technique proposed by \citet{lv2007single}. In their approach, actions are modelled as series of synthetic 2D human poses rendered from many viewpoints and the transition between the synthetic poses is represented by an Action Net graph. However, the rendered images are not realistic as they do not model variations in clothing, background and lighting as in our case. Moreover, our method directly learns features from the rendered images rather than hand crafting features.

Another closely related work to our method is the 3D exemplars for action recognition proposed by \citet{weinland2007action}. In their framework, actions are modelled with 3D occupancy grids built from multiple viewpoints. The learned 3D exemplars are then used to produce 2D images that are compared to the observations during recognition. This method essentially  relies on silhouettes which may not be reliably extracted from the test videos especially under challenging background/lighting conditions.

\citet{Hankelets} proposed Hankelet which is a viewpoint invariant representation that captures the dynamic properties of short tracklets. Hanklets do not carry any spatial information and their viewpoint invariant properties are limited. Very early attempts for view invariant human action recognition include the following methods. \citet{yilmaz2005actions} proposed action sketch, an action representation that is a sequence of the 2D contours of an action in the $x,y,t$ space-time. Such a representation is not completely viewpoint invariant. \citet{parameswaran2006view} used 2D projections of 3D human motion capture data as well on manually segmented real image sequences to perform viewpoint robust action recognition. \citet{rao2002view} used the spatio-temporal 2D trajectory curvatures as a compact representation for view-invariant action recognition. However, the same action can result in very different 2D trajectories when observed from different viewpoints. \citet{weinland2006free} proposed Motion History Volumes (MHV) as a viewpoint invariant representation for human actions. MHVs are aligned and matched using Fourier Transform. This method requires multiple calibrated and background-subtracted video cameras which is only possible in controlled environments. 

View knowledge transfer methods transfer features of different viewpoints to a space where they can be directly matched to achieve viewpoint invariant action recognition. Early methods in this category learned similar features between different viewpoints. For example, \citet{farhadi2008learning} represented actions with histograms of silhouettes and optical flow and  learned features with maximum margin clustering that are similar in different views. Source views are then transferred to the target view before matching. Given sufficient multiview training instances, it was shown later that a hash code with shared values can be learned \citep{farhadi2009latent}. \citet{gopalan2011domain} used domain adaptation for view transfer. \citet{liu2011cross} used a bipartite graph to model two view-dependent vocabularies and applied bipartite graph partitioning to co-cluster two vocabularies into visual-word clusters called bilingual-words that bridge the semantic gap across view-dependent vocabularies. More recently, \citet{DVV} proposed the idea of virtual views that connect action descriptors from one view to those extracted from another view. Virtual views are learned through linear transformations of the action descriptors. \citet{CVP} proposed the idea of continuous virtual path that connects actions from two different views. Points on the virtual path are virtual views obtained by linear transformations of the action descriptors. They proposed a virtual view kernel to compute similarity between two infinite-dimensional features that are concatenations of the virtual view descriptors leading to kernelized classifiers. \citet{zheng2013learning} learned a view-invariant sparse representation for cross-view action recognition. \citet{NKTM} proposed a non-linear knowledge transfer model that mapped dense trajectory action descriptors to canonical views. However, this method does not exploit the appearance/shape features.

Deep learning has also been used for action recognition. \citet{simonyan2014two} and \citet{twoStream_CVPR16} proposed two stream CNN architectures using appearance and optical flow to perform action recognition.\citet{wang2015action} proposed a two stream structure to combine hand-crafted features and deep learned features. They used trajectory pooling for one stream and deep learning for the second and combined the features from the two streams to form trajectory-pooled deep-convolutional descriptors. Nevertheless, the method did not explicitly address viewpoint variations. \citet{toshev2014deeppose} proposed DeepPose for human pose estimation based on Deep Neural Networks which treated pose estimation as a regression problem and represented the human pose body joint locations. This method is able to capture the context and reasoning about the pose in a holistic manner however, the scale of the dataset used for training was limited and the method does not address viewpoint variations. \citet{pfister2015flowing} proposed a CNN architecture to estimate human poses. Their architecture directly regresses pose heat maps and combines them with optical flow. This architecture relies on neighbouring frames for pose estimations. \citet{C3D_pami} proposed a 3D Convolutional Neural Network (C3D) for human action recognition. They used a set of hard-wired kernels to generate multiple information channels corresponding to the gray pixel values, ($x,y$) gradients, and ($x,y$) optical flow from seven input frames. This was followed by three convolution layers whose parameters were learned through back propagation. The C3D model did not explicitly address invariance to viewpoint or other factors.

\citet{wang2016action} proposed joint trajectory maps, projections of 3D skeleton sequences to multiple 2D images, for human action recognition. \citet{karpathy2014large} suggested a multi-resolution foveated architecture for speeding up CNN training for action recognition in large scale videos. \citet{varol2017long} proposed long-term temporal convolutions (LTC) and showed that LTC-CNN models with increased temporal extents improve action recognition accuracy. \citet{tran2015learning} treated videos as cubes and performed convolutions and pooling with 3D kernels. Recent methods \citep{li2016vlad3, zhu2016key, wang2016improving, zhang2016real, su2016hierarchical, wang2016temporal} emphasize on action recognition in large scale videos where the background context is also taken into account. 

\citet{shahroudy2016multimodal} divided the actions into body parts and proposed a multimodal-multipart learning method to represent their dynamics and appearances. They selected the discriminative body parts by integrating a part selection process into the learning and proposed a hierarchical mixed norm to apply sparsity between the parts, for group feature selection. This method is based on depth and skeleton data and uses LOP (local occupancy patterns) and HON4D (histogram of oriented 4D normals) as features. \citet{yu2016structure} proposed a Structure Preserving Projection (SPP) to represent RGB-D video data fusion. They described the gradient fields of RGB and depth data with a new Local Flux Feature (LFF), and then fused the LFFs from RGB and depth channels. With structure-preserving projection, the pairwise structure and bipartite graph structure are preserved when fusing RGB and depth information into a Hamming space, which benefits the general action recognition. 

\citet{huang2016deep} incorporated the Lie group structure into deep learning, to transform  high-dimensional Lie group trajectory into temporally aligned Lie group features for skeleton-based action recognition. The incorporated learning structure generalizes the traditional neural network model to non-Euclidean Lie groups. \citet{luo2017unsupervised} proposed to use Recurrent Neural Network based Encoder-Decoder framework to learn video representation in capturing motion dependencies. The learning process is unsupervised and it focuses on encoding the sequence of atomic 3D flows in consecutive frames.

\citet{jia2014latent} proposed a latent tensor transfer learning method to transfer knowledge from the source RGB-D dataset to the target RGB only dataset such that the missing depth information in the target dataset can be compensated. The learned 3D geometric information is then coupled with RGB data in a cross-modality regularization framework to align them. However, to learn the latent depth information for RGB data, a RGB-D source dataset is required to perform the transfer learning, and for different source datasets, the learned information may not be consistent which could affect the final performance. \citet{kong2017max} proposed max-margin heterogeneous information machine (MMHIM) to fuse RGB and depth features. The histograms of oriented gradients (HOG) and histograms of optical flow (HOF) descriptors are projected into independent shared and private feature spaces, and the features are represented in matrix forms to build a low-rank bilinear model for the classification. This method utilizes the cross-modality and private information, which are also de-noised before the final classification.

\citet{kerola2017cross} used spatio-temporal key points (STKP) and skeletons to represent an action as a temporal sequence of graphs, and then applied the spectral graph wavelet transform to create the action descriptors. \citet{varol17} recently proposed SURREAL to synthesize human pose images for the task of body segmentation and depth estimation. The generated dataset includes RGB images, together with depth maps and body parts segmentation information. They then learned a CNN model from the synthetic dataset, and then conduct pixel-wise classification for the real RGB pose images. This method made efforts in diversifying the synthetic data, however, it didn't address the distribution gap between synthetic and real images. Moreover, this method only performs human body segmentation and depth estimation.

Our survey shows that none of the existing techniques explicitly learn invariant features through a training dataset that varies all irrelevant factors for the same human pose. This is partly because such training data is very difficult and expensive to generate. 
We resolve  this problem by developing a method to generate such data synthetically.  The proposed data generation technique is a major contribution of this work that can  synthesize large amount of data for training data-hungry deep  network models. By easily introducing a variety of action irrelevant  variations in the synthetic data, it is possible to learn effective models that can extract invariant information. In this work, we propose Human Pose Models for extracting such information from both RGB and depth images.

\begin{figure}[t]
\centering
\includegraphics[width=\textwidth]{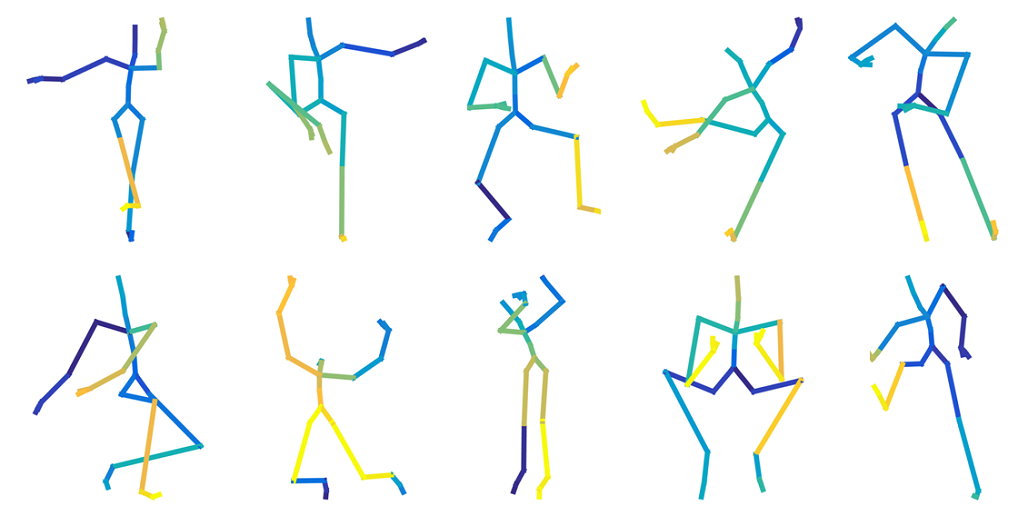}
\caption{Samples from the 339 representative human poses learned from the CMU MoCap skeleton data}
\label{fig:pose_dictionary}
\vspace{-3mm}
\end{figure}

\vspace{-2mm}
\section{Generating Synthetic Training Data}
\label{sec:data}
The proposed synthetic data generation steps are explained in the following subsections. Unless specified, the steps are shared by RGB and depth data generation.

\vspace{-2mm}
\subsection{Learning Representative Human Poses}

Since the space of possible human poses is extremely large, we learn a finite number of representative human poses in a way that is not biased by irrelevant factors such as body shapes, appearances, camera viewpoints, illumination and backgrounds. Therefore, we learn the representative poses from 3D human joints (skeletons), because each skeleton can be fitted with 3D human models of any size/shape. The CMU MoCap database is ideal for this purpose because it contains joint locations of real humans performing different actions resulting in a large number of different poses. This data consists of over 2500 motion sequences and over 200,000 human poses. We randomly sample 50,000 frames as the pose candidates and cluster them with HDBSCAN algorithm \citep{mcinnes2017hdbscan} using the skeletal distance function \citep{shakhnarovich2005learning}  
\begin{equation}
D(\theta_1, \theta_2) = \max_{i \leq j \leq L} \sum_{d \in x,y,z} \mid \theta_{d,1}^i - \theta_{d,2}^j \mid
\label{eq:1}
\end{equation} 
where $\theta_1$ and $\theta_2$ are the $x,y,z$ joint locations of two skeletons. By setting the minimum cluster size to 20, the HDBSCAN algorithm outputs 339 clusters and we choose the pose with highest HDBSCAN score in each cluster to form the representative human poses. Figure \ref{fig:pose_dictionary} shows a few of the learned representative human poses.

\begin{figure}[t]
\centering
\includegraphics[width=\textwidth]{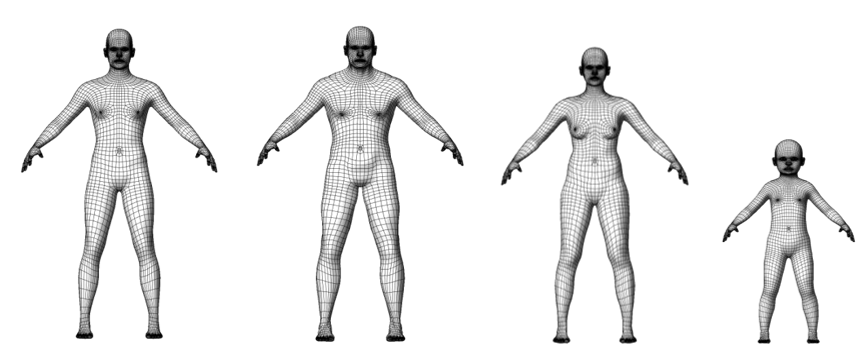}
\caption{3D human (male, heavy male, female, child) models generated with the MakeHuman software}
\label{fig:3D_humans}
\end{figure}

\vspace{-2mm}
\subsection{Generating 3D Human Models}
The 339 representative human pose skeletons are fitted with full 3D human models. We use the open source MakeHuman\footnote{http://www.makehuman.org} software to generate 3D human models because it has three attractive properties. Firstly, the 3D human models created by MakeHuman contain information for fitting the model to the MoCap skeletons to adopt that pose. Secondly, it is possible to vary the body shape, proportion and gender properties to model different shape humans. Figure \ref{fig:3D_humans} shows the four human body shapes we used in our experiments. Thirdly, MakeHuman allows for selecting some common clothing types as shown in Fig. \ref{fig:default_clothing}. Although, the MakeHuman offers limited textures for the clothing, we write a Python script to apply many different types of clothing textures obtained from the Internet.

\subsection{Fitting 3D Human Models to MoCap Skeletons}
\begin{figure}[t]
\centering
\includegraphics[width=\textwidth]{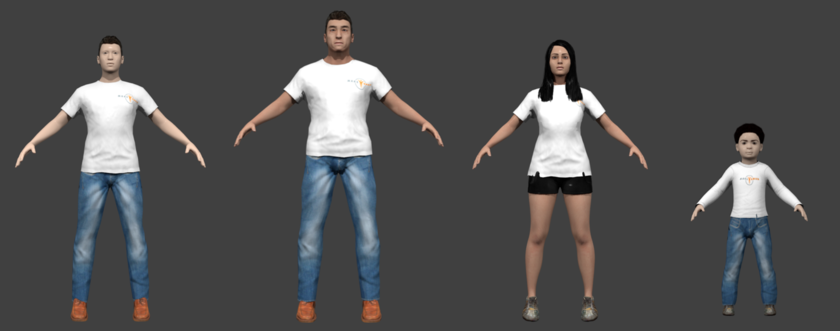}
\caption{Sample clothing given by the MakeHuman software to the 3D models}
\label{fig:default_clothing}
\vspace{-1mm}
\end{figure}

MakeHuman generates 3D human models in the same canonical pose as shown in Figure \ref{fig:3D_humans} and \ref{fig:default_clothing}. We use the open source Blender\footnote{http://www.blender.org} software to fit the 3D human models to the 339 representative human pose skeletons. Blender loads the 3D human model and re-targets its rigs to the selected MoCap skeleton. As a result, the 3D human model adopts the pose of the skeleton and we get the representative human poses as full 3D human models with different clothing types and body shapes. Clothing textures are varied later.

\subsection{Multiview Rendering to Generate RGB Images}

We place each 3D human model (with a representative pose) in different backgrounds and lighting conditions using Blender. In the following, we explain how different types of variations were introduced in the rendered images.

\begin{figure}[t]
\centering
\includegraphics[width=\textwidth]{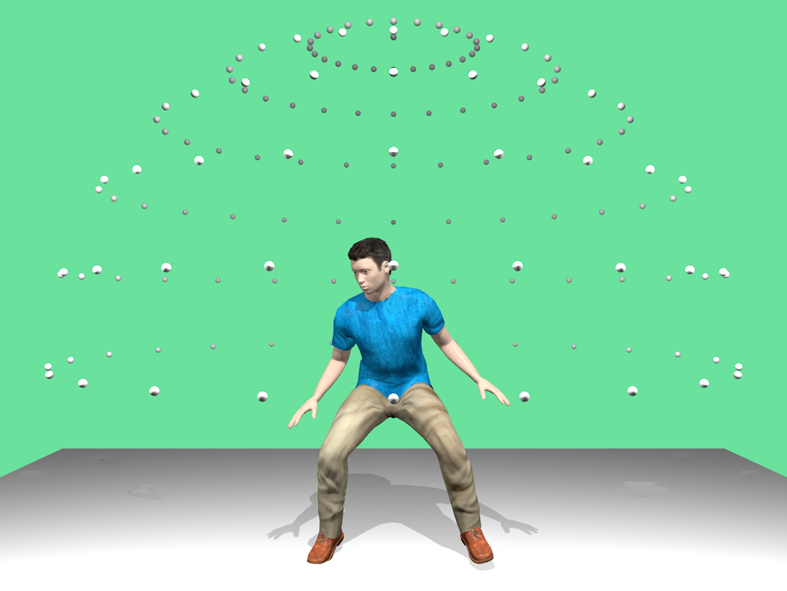}
\caption{Each 3D human model is rendered from 180 camera viewpoints on a hemisphere. All cameras point to the center of the human model}
\label{fig:mv_render}
\vspace{-1mm}
\end{figure}

\begin{figure}[t]
\centering
\includegraphics[width=\textwidth]{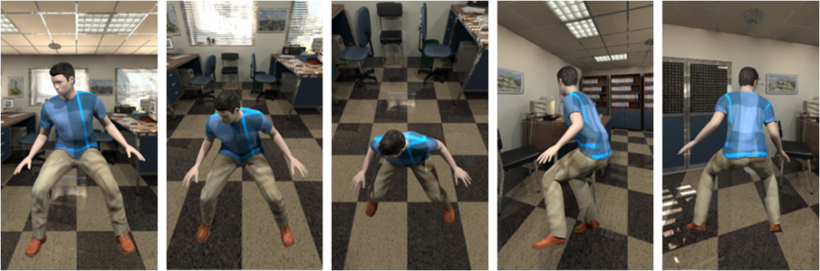}
\caption{Human pose images rendered from multiple viewpoints}
\label{fig:view_variance}
\vspace{-1mm}
\end{figure}

\vspace{2mm}
\noindent {\bf Camera Viewpoint:}   We place 180 virtual cameras on a hemisphere over the 3D human model to render RGB images. These cameras are 12 degrees apart along the latitude and longitude and each camera points to the center of the 3D human model. Figure \ref{fig:mv_render} illustrates the virtual cameras positioned around a 3D human model where no background has been added yet to make the cameras obvious. Figure \ref{fig:view_variance} shows a few images rendered from multiple viewpoints after adding the background and lighting.

\vspace{2mm}
\noindent {\bf Background and Lighting:} We incorporate additional rich appearance variations in the background and lighting conditions to synthesize images that are as realistic as possible. Background variation is performed in two modes. One, we download thousands of 2D indoor scenes from Google Images and randomly select one as Blender background during image rendering. Two,  we download 360$^{\rm o}$ spherical High Dynamic Range Images (HDRI) from Google Images and use them as the environmental background in Blender. In the latter case, when rendering images from different viewpoints, the background changes accordingly. Figure \ref{fig:view_variance} shows some illustrations. In total, we use 2000 different backgrounds that are mostly indoor scenes, building lobbies with natural lighting and a few outdoor natural scenes. We place three lamps at different locations in the scene and randomly change their energy to achieve lighting variations.

\begin{figure}[t]
\centering
\includegraphics[width=\textwidth]{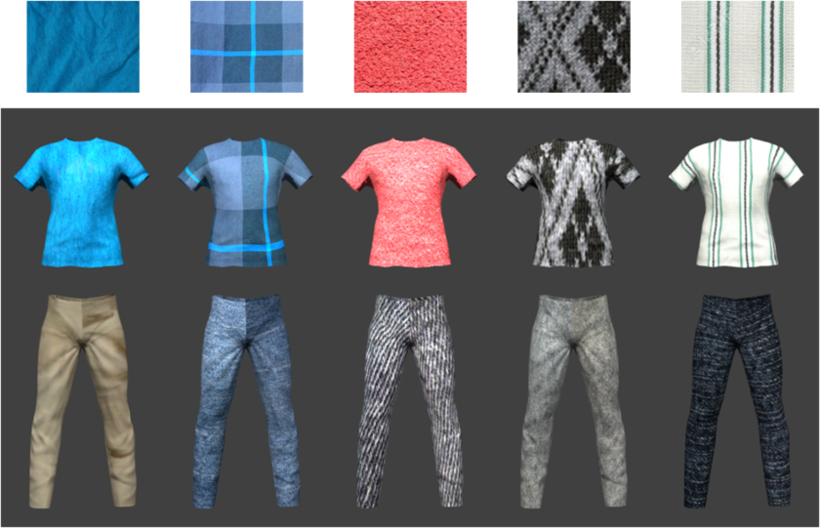}
\caption{We use 262 textures for shirts and 183 for the trousers/shorts. The clothing textures of the human models are varied randomly through a Python script in Blender}
\label{fig:clothing}
\end{figure}

\begin{figure}[t]
\centering
\includegraphics[width=\textwidth]{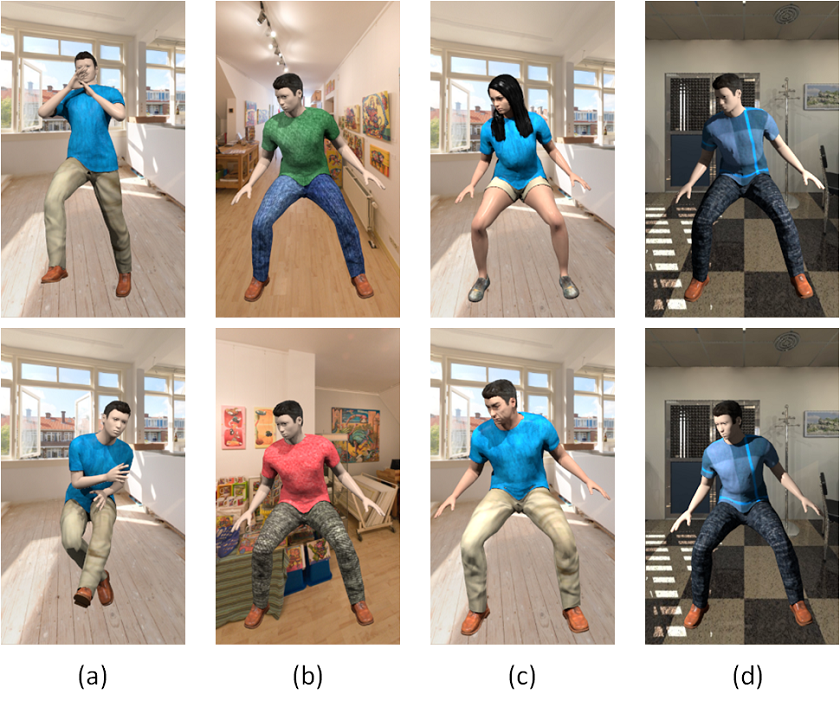}
\caption{Synthetic RGB images where variations are introduced in (a) human pose, (b) background and clothing texture, (c) body shapes and (d) lighting conditions}
\label{fig:all_variances}
\end{figure}

\vspace{2mm}
\noindent {\bf Clothing Texture:} Clothing texture is varied by assigning different textures, downloaded from Google Images, to the clothing of the 3D human models. In total, we used 262 different textures of shirts and 183 textures for trousers/shorts to generate our training data. Figure \ref{fig:clothing} shows some of the clothing textures we used and Figure \ref{fig:all_variances} shows some rendered images containing all types of variations.

\subsection{Multiview Rendering to Generate Depth Images}

\begin{figure}[t]
\centering
\includegraphics[width=\textwidth]{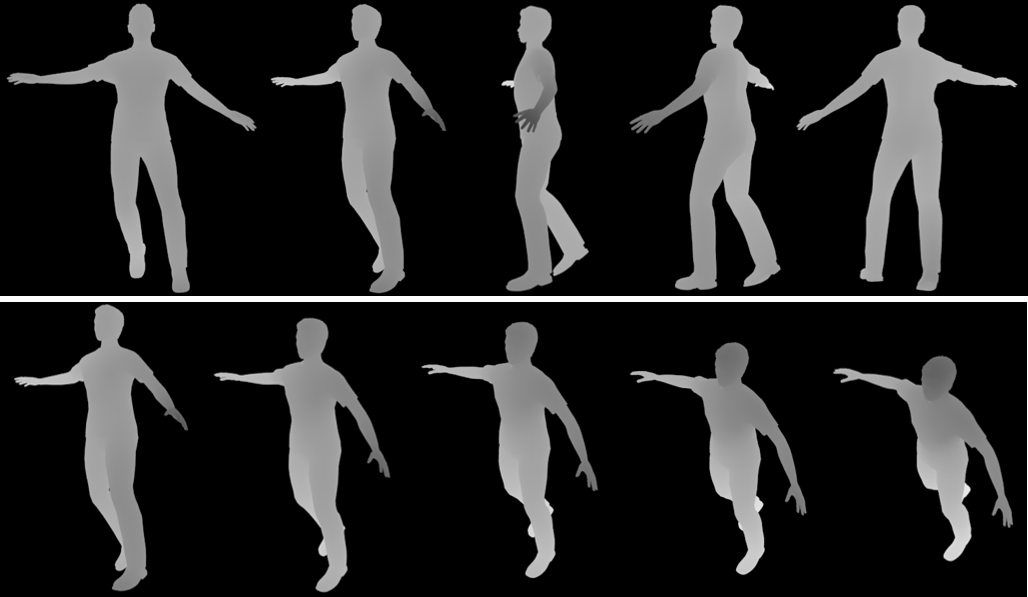}
\caption{Depth human pose images rendered from multiple viewpoints. The first row of the figure illustrates horizontal viewpoint change, and the second row illustrates vertical viewpoint change}
\label{fig:depth_mv_samples}
\vspace{-1mm}
\end{figure}

Depth images simply record the distance of the human from the camera without any background, texture or lighting variation. Therefore, we only vary the camera viewpoint, clothing types (not textures) and body shapes when synthesizing depth images. The virtual cameras are deployed in a similar way to the RGB image rendering. Figure \ref{fig:depth_mv_samples} shows some depth images rendered from different viewpoints. In the Blender rendering environment, the bounding box of the human model is recorded as a group of vertices, which can be converted to an $xy$ bounding box around the human in the rendered image  coordinates. This bounding box is used to crop the human in the rendered depth images as well as RGB images. The cropped images are used to learn the Human Pose Models.

\subsection{Efficiency of Synthetic Data Generation}

To automate the data generation, we implemented a Python script\footnote{The data synthesis script will be made public.} in Blender on a 3.4GHz machine with 32GB RAM. The script runs six separate processing threads and on the average, generates six synthetic pose images per second. Note that this is an off-line process and can be further parallelized since each image is rendered independently. Moreover, the script can be implemented on the cloud for efficiency without the need to upload any training data. The training data is only required for learning the model and can be deleted afterwards. For each of the 339 representative poses, we generate images from all 180 viewpoints while applying a random set of other variations (clothing, background, body shape and lighting). In total, about 700,000 synthetic RGB and depth images are generated to train the proposed HPM$_{\mathrm{RGB}}$ and HPM$_{\mathrm{3D}}$.

\section{Synthetic Data Refinement with GANs}
\label{sec:GAN}

The synthetic images are labelled with 339 different human poses and cover common variations that occur in real data. They can be used to learn Human Pose Models that transfer human poses to a high level view-invariant space. However, it is likely that the synthetic images are sampled from a distribution which is different from the distribution of real images. Given this distribution gap, the Human Pose Models learned from synthetic images may not generalize well to real images. Therefore, before learning the models, we minimize the gap between the distributions. For this purpose, we adopt the simulated and unsupervised learning framework (SimGAN) \citep{shrivastava2016learning} proposed by Shrivastava \etal. This framework uses an adversarial network structure similar to the Generative Adversarial Network (GAN) \citep{goodfellow2014generative}, but the learning process is based on synthetic images, instead of random noises as in the original GAN method. The SimGAN framework learns two competing networks, refiner $R_\theta(x)$ and discriminator $D_\phi(\tilde{x},y)$, where $x$ is synthetic image, $y$ is unlabelled real image, $\tilde{x}=R_\theta(x)$ is refined image. The loss function of these two networks are defined as $\mathcal{L}_R(\theta)$ and $\mathcal{L}_D(\phi)$ \citep{shrivastava2016learning}
\begin{equation} 
    \mathcal{L}_R(\theta)\!\! = \! -\!\!\sum_{i}log(1-D_\phi(R_\theta(x_i))) + \lambda\|R_\theta(x_i)-x_i\|_1,
 \end{equation}
 \vspace{-5mm}
 \begin{equation} 
    \mathcal{L}_D(\phi) = -\sum_{i}log(D_\phi(\tilde{x_i})) - \sum_{j}log(1-D_\phi(y_j)),
\end{equation}
where $x_i$ is the $i^{th}$ synthetic image, $\tilde{x}_i$ is its corresponding refined image, $y_j$ is the $j^{th}$ real image, $\|.\|_1$ is $\ell_1$ norm, and $\lambda\in[0,1]$ is the regularization factor.

\subsection{Implementation Details}

We modify the Tensorflow implementation\footnote{https://github.com/carpedm20/} of SimGAN to make it suitable for our synthetic RGB images. For the refiner network, we extend the input data channel from 1 to 3. The input images are first convolved with 7 $\times$ 7 filters and then converted into 64 feature maps. The 64-channel feature maps are passed through multiple ResNet blocks. The setting of ResNet blocks and the structure of discriminator network are the same as \citep{shrivastava2016learning}.

To get benchmark distribution for the synthetic images, we randomly select 100,000 unlabelled real images from the NTU RGB+D Human Activity Dataset \citep{shahroudy2016ntu}. Each image is cropped to get the human body as the region of interest and then resized to 224 $\times$ 224. Through adversarial learning, SimGAN framework \citep{shrivastava2016learning} will force the distribution of synthetic images to approach this benchmark distribution. Although, we use samples from the NTU RGB+D Human Activity Dataset as benchmark to train the SimGAN network, this is not mandatory as any other dataset containing real human images can be used. This is because the SimGAN learning is an unsupervised process, which means no action labels are required. Our experiments in later sections also illustrate that the performance improvement gained from GAN-refinement has no dependence on the type of real images used for SimGAN learning.

\subsection{Qualitative Analysis of GAN Refined Images} 
\label{section:GAN_analysis}

We compare the real images, raw synthetic images, and GAN-refined synthetic images, to analyse the effect of GAN refinement on our synthetic RGB and depth human pose datasets.

\begin{figure}[t]
\centering
\includegraphics[width=\textwidth]{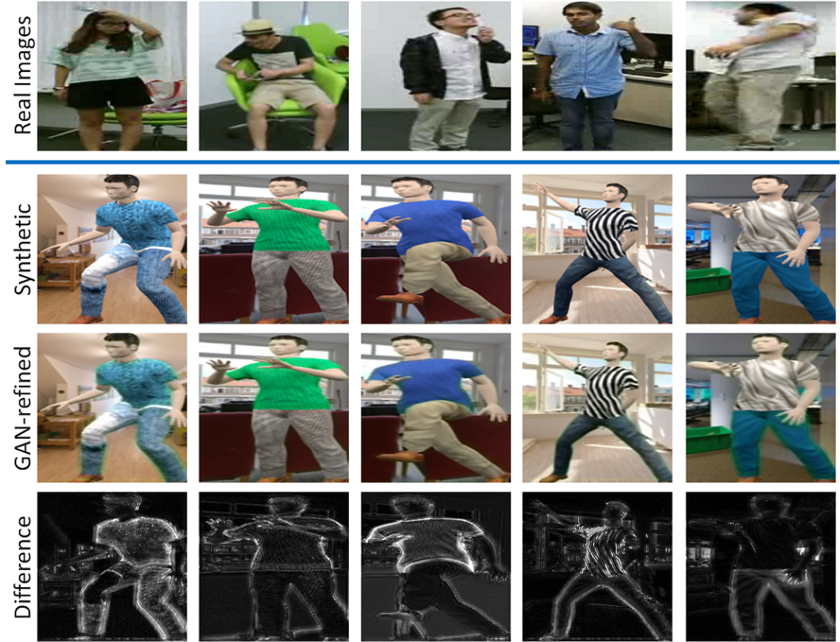}
\caption{Comparing real, raw synthetic and GAN-refined synthetic RGB images. Last row shows the different between raw and GAN-refined synthetic images. Since the backgrounds used are already real, differences are mostly on the synthetic human body especially at their edges}
\label{fig:rgb_GAN_compare}
\vspace{-1mm}
\end{figure}

Figure~\ref{fig:rgb_GAN_compare} compares the real, raw synthetic and GAN-refined synthetic RGB images. One obvious difference between real and synthetic RGB images is that the synthetic ones are sharper and have more detail than the real images. This is because the synthetic images are generated under ideal conditions and the absence of realistic image noises makes them different from real images. However, with GAN learning, the refined RGB images lose some of the details (i.e. they are not as sharp) and become more realistic. The last row of Figure \ref{fig:rgb_GAN_compare} shows the difference between the raw and refined synthetic images. Notice that the major differences (bright pixels) are at the locations of the humans and especially at their boundaries whereas the backgrounds have minimal differences (dark pixels). The reason for this is that the synthetic images are created using synthetic 3D human models but real background images i.e. rather than building Blender scene models (walls, floors, furnitures, etc.) from scratch, we used real background images for efficiency and diversity of data. Moreover, the Blender lighting function causes shading variation on the human models only, and the shading effects of backgrounds always remain the same. All these factors make the human model stand out of the background. On the other hand, the human subjects are perfectly blended with the background in the real images. The GAN refinement removes such differences in synthetic images and makes the human models blend into the background. Especially, the bright human boundaries (last row) shows that the GAN refinement process is able to sense and remove the difference between human model and background images.

\begin{figure}[t]
\includegraphics[width=\textwidth]{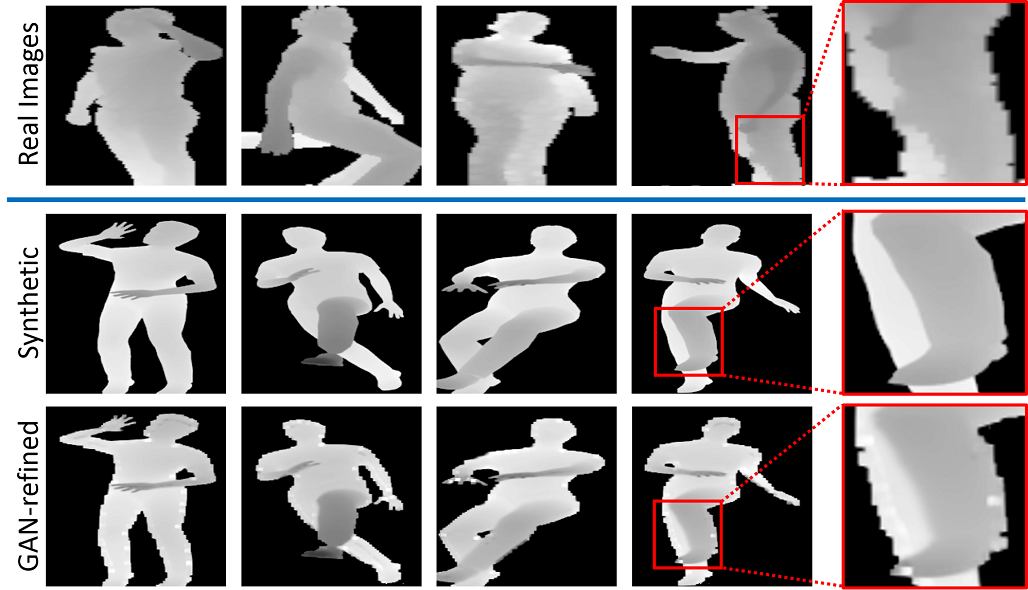}
\caption{Comparing real, raw synthetic and GAN-refined synthetic depth images}
\label{fig:depth_GAN_compare}
\end{figure}

Figure~\ref{fig:depth_GAN_compare} shows a similar comparison for depth images. The most obvious difference between real and synthetic depth images is the noise along the boundary. The edges in the real depth images are not smooth whereas they are very smooth in the synthetic depth images. Other differences are not so obvious to the naked eye but nevertheless, these differences might limit the generalization ability of the Human Pose Model learned from synthetic images. The third row of Fig.~\ref{fig:depth_GAN_compare} shows the refined synthetic depth images and the last column shows the difference between the images in more detail. Thus the GAN refinement successfully learns to model boundary noise and other non-obvious differences of real images and applies them to the synthetic depth maps narrowing down their distribution gap.

In the experiments, we will show quantitative results indicating that the Human Pose Models learned from the GAN refined synthetic images outperform those that are learned from raw synthetic images.

\section{Learning the Human Pose Models}

Every image in our synthetic data has a label corresponding to one of the 339 representative human poses. For a given human pose, the label remains the same irrespective of the camera viewpoint, clothing texture, body shape, background and lighting conditions. We learn CNN models that map the rendered images to their respective human pose labels. We learn HPM$_{\mathrm{RGB}}$ and HPM$_{\mathrm{3D}}$ for RGB and depth images independently and test three popular CNN architectures, i.e. AlexNet \citep{AlexNet}, GoogLeNet \citep{GoogLeNet}, and ResNet-50 \citep{ResNet}, to find the most optimal architecture through controlled experiments. These CNN architectures performed well in the ImageNet Large Scale Visual Recognition Challenge (ILSVRC) in 2012, 2014, and 2015 respectively, and come with increasing number of layers. We fine tune the ILSVRC pre-trained models using our synthetic data and compare their performance on human action recognition.

\subsection{Model Learning}

The three pre-trained models have a last InnerProduct layer with 1000 neurons. For fine tuning, we replace the last layer with a 339 neuron layer representing the number of classes in our synthetic human pose dataset. All synthetic images are cropped to include only the human body and then resized to 256$\times$256 pixels. During training, these images are re-cropped to the required input dimension for the specific network with default data augmentation, and are also mirrored with a probability of 0.5. We use the synthetic pose images from 162 randomly selected camera viewpoints for training, and the images from the remaining 18 cameras for validation.

The Caffe library \citep{jia2014caffe} is used to learn the proposed HPM$_{\mathrm{RGB}}$ and HPM$_{\mathrm{3D}}$ models. The initial learning rate of the model was set to 0.01 for the last fully-connected layers and 0.001 for all other layers. We used a batch size of 100 and trained the model for 3 epochs. We decreased the learning rate by a factor of 10 after every epoch. Training was done using a single NVIDIA Tesla K-40 GPU.

\subsection{Extracting Features from Real Videos}

To extract features from real videos, the region containing the human is first cropped from each frame and then the cropped region is resized to match the input of the network. The cropped-resized regions from each frame are passed individually through the learned Human Pose Model (HPM$_{\mathrm{RGB}}$ for RGB frames, and HPM$_{\mathrm{3D}}$ for depth frames), and a layer prior to the labels is used as invariant representation of the human pose. Specifically, we use fc7 layer for AlexNet, pool5/7x7$\_$s1 layer for GoogLeNet and pool5 for ResNet-50. This representation is unique, compact, invariant to the irrelevant factors and  has the added advantage that it aligns the features between multiple images. While the pixels of one image may not correspond to the pixels of another image, the individual variables of the CNN features are aligned. Therefore, we can perform temporal analysis along the individual variables.

\section{Temporal Representation and Classification}
For temporal representation, we use the Fourier Temporal Pyramid (FTP) \citep{wang2013learning} on the features extracted from the video frames. Temporal representation for HPM$_{\mathrm{3D}}$ and HPM$_{\mathrm{RGB}}$ features is done in a similar way and explained in general in the next paragraph.

Let $V_t^i$ denote the $t$-th frame of $i$-th video, $t=1,2,\dots,f$ where $f$ is the total number of frames. Take HPM with GoogleNet structure as an example, denote the pool5/7x7$\_$s1 layer activations of frame \(V_t^i \) as \(A_t^i \in \mathbb{R}^{1024 \times 1} \) and the frame-wise pose features of the $i$-th video as \(A^i=[A_1^i, A_2^i, \dots, A_f^i]^T\). FTP is applied on \(A^i\) for temporal encoding using a pyramid of three levels where \(A^i\) is divided in half at each level giving $1+2+4=7$ feature groups. Short Fourier Transform is applied to each feature group, and the first four low-frequency coefficients (i.e. $4 \times 7 = 28$) are used to form a spatio-temporal action descriptor $B^i \in \mathbb{R}^{1024 \times 28}$. Finally, $B^i$ is stretched to $D^i \in \mathbb{R}^{1 \times 28672}$ to get the final spatio-temporal representation of the $i$-th video. When the dimension of frame-wise pose feature changes, the dimension of spatio-temporal descriptor changes accordingly, for example, $B^i \in \mathbb{R}^{4096 \times 28}$ for AlexNet, and $B^i \in \mathbb{R}^{2048 \times 28}$ for ResNet-50.

Note that the FTP encodes the temporal variations of the RGB action videos in the HPM$_{\mathrm{RGB}}$ feature space. The video frames are first aligned in the $A_t^i$ HPM$_{\mathrm{RGB}}$ feature space which makes it possible to preserve the spatial location of the features while temporal encoding with FTP. On the other hand, dense trajectories model temporal variations in the pixel space (of RGB videos) where pixels corresponding to the human body pose are not aligned. This is the main reason why dense trajectory features are encoded with Bag of Visual Words (BoVW) which facilitates direct matching of dense trajectory features from two videos. However, this process discards the spacial locations of the trajectories. Thus, similar trajectories from different locations in the frame will vote to the same bin in BoVW feature.

An advantage of performing temporal encoding in different feature spaces is that the features are non-redundant. Thus our HPM$_{\mathrm{RGB}}$ and dense trajectories capture complementary information. Although dense trajectories cannot capture the appearance information, they are somewhat robust to viewpoint changes as shown in \citep{rahmani2017learning}. Therefore, we augment our HPM$_{\mathrm{RGB}}$ features with dense trajectory features before performing classification. We use the improved dense trajectories (iDT) \citep{wang2013action} implementation which provides additional features such as HOG \citep{dalal2005histograms}, HOF and MBH \citep{dalal2006human}. However, we only use the trajectory part and discard HOG/HOF/MBH features for two reasons. Firstly, unlike our HPMs, HOG/HOF/MBH features are not view-invariant. Secondly, our HPMs already encode the appearance information. We use the NKTM \citep{NKTM} codebook to encode the trajectory features and denote the encoded BoVW as $D^i \in \mathbb{R}^{2000}$ for video $V^i$.

We use SVM \citep{REF08a} for classification and report results in three settings i.e. RGB, depth and RGB-D. In the RGB setting, we represent the HPM$_{\mathrm{RGB}}$ features temporally encoded with FTP and then combine them with the trajectory BoVW features since both types of features can be extracted from RGB videos. In the depth setting, we represent the HPM$_{\mathrm{3D}}$ features with FTP but do not combine trajectory features because trajectories cannot be reliably extracted from the depth videos. In the RGB-D setting, we combine the FTP features from both HPM models with the trajectory BoVW features.

\begin{figure}[t]
\centering
\includegraphics[width=\textwidth]{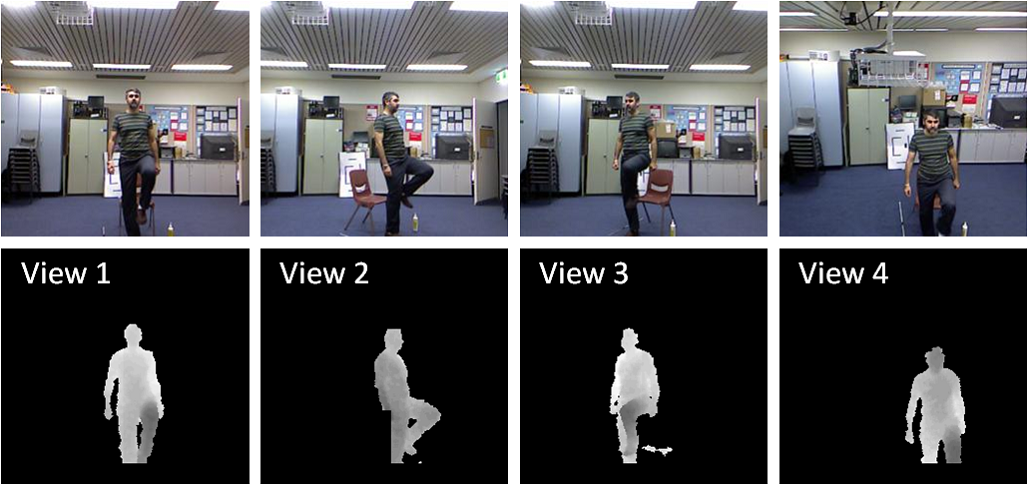}
\caption{Sample frames from the UWA3D Multiview Activity II dataset \citep{HOPC2016PAMI}}
\label{fig:uwa3d_sample}
\end{figure}

\section{Datasets}

Experiments are performed in the following three benchmark datasets for cross-view human action recognition.

\subsection{UWA3D Multiview Activity-II Dataset}

Figure \ref{fig:uwa3d_sample} shows sample frames from this dataset. The dataset \citep{HOPC2016PAMI} consists of 30 human actions performed by 10 subjects and recorded from 4 different viewpoints at different times using the Kinect v1 sensor. The 30 actions are: (1) one hand waving, (2) one hand punching, (3) two hands waving, (4) two hands punching, (5) sitting down, (6) standing up, (7) vibrating, (8) falling down, (9) holding chest, (10) holding head, (11) holding back, (12) walking, (13) irregular walking, (14) lying down, (15) turning around, (16) drinking, (17) phone answering, (18) bending, (19) jumping jack, (20) running, (21) picking up, (22) putting down, (23) kicking, (24) jumping, (25) dancing, (26) moping floor, (27) sneezing, (28) sitting down (chair), (29) squatting, and (30) coughing. The four viewpoints are: (a) front, (b) left, (c) right, (d) top.

This dataset is challenging because of the large number of action classes and because the actions are not recorded simultaneously leading to intra-action differences besides viewpoint variations. The dataset also contains self-occlusions and human-object interactions in some videos. 

We follow the protocol of \citep{HOPC2016PAMI} where videos from two views are used for training and the videos from the remaining views are individually used for testing leading to 12 different cross-view combinations in this evaluation protocol.

\subsection{Northwestern-UCLA Multiview Dataset}
\begin{figure}[t]
\centering
\includegraphics[width=\textwidth]{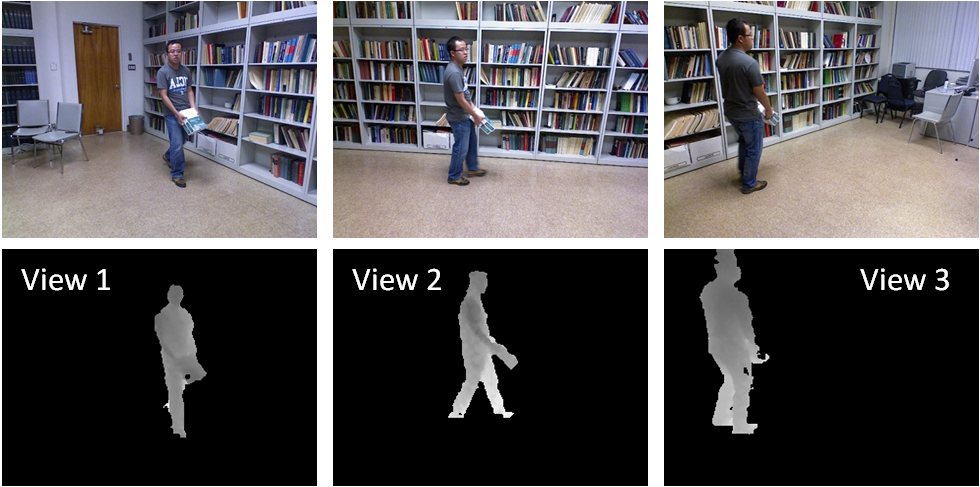}
\caption{Sample frames from the Northwestern-UCLA Multiview Action dataset \citep{AOG}}
\label{fig:nucla_sample}
\end{figure}

This dataset \citep{AOG} contains RGB-D videos captured simultaneously from three different viewpoints with the Kinect v1 sensor. Figure \ref{fig:nucla_sample} shows sample frames of this dataset from the three viewpoints. The dataset contains RGB-D videos of 10 subjects performing 10 actions: (1) pick up with one hand, (2) pick up with two hands, (3) drop trash, (4) walk around, (5) sit down, (6) stand up, (7) donning, (8) doffing, (9) throw, and (10) carry. The three viewpoints are: (a) left, (b) front, and (c) right.  This dataset is very challenging because many actions share the same ``walking'' pattern before and after the actual action is performed. Moreover, some actions such as ``pick up with on hand'' and ``pick up with two hands'' are hard to distinguish from different viewpoints. 

We use videos captured from two views for training and the third view for testing making three possible cross-view combinations.

\subsection{NTU RGB+D Human Activity Dataset}

\begin{figure}[t]
\centering
\includegraphics[width=\textwidth]{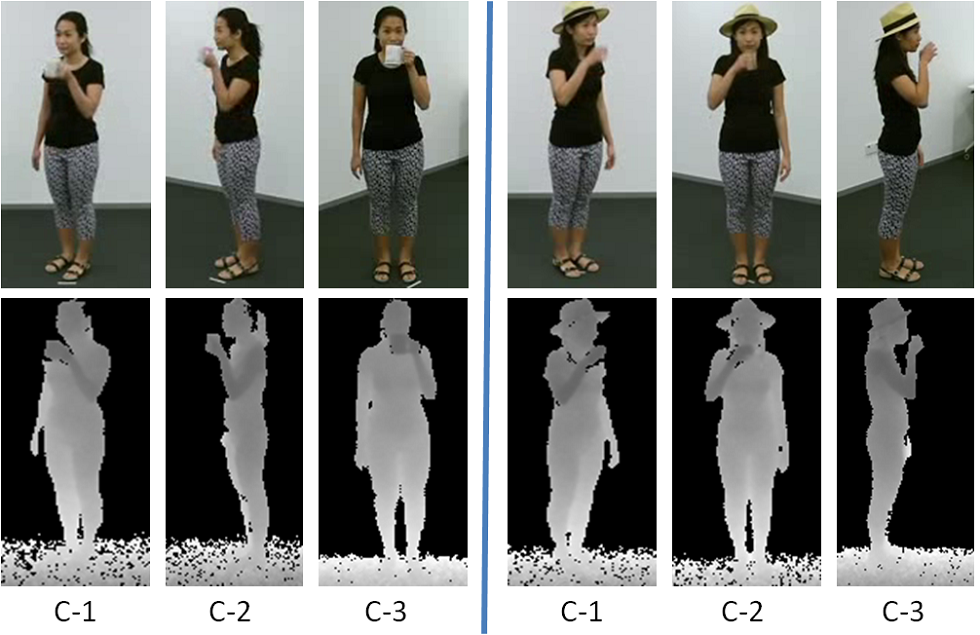}
\caption{RGB and depth sample frames from the NTU RGB+D Human Activity Dataset \citep{shahroudy2016ntu}. Three sensors C-1, C-2 and C-3 are used to record this dataset. The left group of images in this figure shows the actions recorded with the performer facing the sensor C-3, and the right group of images are recorded when the action performer faces the sensor C-2}
\label{fig:ntu_samples}
\end{figure}

The NTU RGB+D Human Activity Dataset \citep{shahroudy2016ntu} is a large-scale RGB+D dataset for human activity analysis. This dataset was collected with the Kinect v2 sensor and includes 56,880 action samples each for RGB videos, depth videos, skeleton sequences and infra-red videos. We only use the RGB and depth parts of the dataset. There are 40 human subjects performing 60 types of actions including 50 single person actions and 10 two-person interactions. Three sensors were used to capture data simultaneously from three horizontal angles: $-45^\circ, 0^\circ, 45^\circ$, and every action performer performed the action twice, facing the left or right sensor respectively. Moreover, the height of sensors and their distance to the action performer were further adjusted to get more viewpoint variations. The NTU RGB+D dataset is the largest and most complex cross-view action dataset of its kind to date. Figure \ref{fig:ntu_samples} shows RGB and depth sample frames in NTU RGB+D dataset.

We follow the standard evaluation protocol proposed in \citep{shahroudy2016ntu}, which includes cross-subject and cross-view evaluations. For cross-subject protocol, 40 subjects are split into training and testing groups, and each group consists of 20 subjects. For cross-view protocol, the videos captured by sensor C-2 and C-3 are used as training samples, and the videos captured by sensor C-1 are used as testing samples. 

\vspace{-2mm}
\section{Experiments}
\label{sec:Exp}




We first use raw synthetic images to train HPM$_{\mathrm{RGB}}$ and HPM$_{\mathrm{3D}}$ for the three different CNN architectures (AlexNet, GoogLeNet, and ResNet-50), and compare their performance on the UWA and NUCLA datasets. The best performing  architecture is then selected and re-trained on the GAN-refined synthetic images. Next, we compare the HPM models trained on raw synthetic images to those trained on the GAN refined synthetic images. Finally, we perform comprehensive experiments to compare our proposed models trained on GAN refined synthetic images to existing methods on all three datasets.

\vspace{-2mm}
\subsection{HPM Performance with Different Architectures}

\begin{table}[t]
\centering
\caption{Comparison of feature dimensionality and action recognition accuracy(\%) for HPM$_{\mathrm{RGB}}$ and HPM$_{\mathrm{3D}}$ trained using different network architectures}
\label{tab:models_compare}
\begin{tabular}{lcccc}
\hline\noalign{\smallskip}
Network & \multicolumn{1}{l}{Layer} & \multicolumn{1}{l}{Dimension} & \multicolumn{1}{l}{HPM$_{\mathrm{RGB}}$} & \multicolumn{1}{l}{HPM$_{\mathrm{3D}}$} \\

\noalign{\smallskip}\hline\noalign{\smallskip}
\multicolumn{ 5}{c}{\bf{UWA3D Multiview Activity-II}} \\
\noalign{\smallskip}\hline\noalign{\smallskip}

AlexNet & fc7 & 4096 & 61.2 & 72.1 \\ 
ResNet-50 & pool5 & 2048 & \bf{65.4} & 74.0 \\
GoogLeNet & pool5 & \bf{1024} & 64.7 & \bf{74.1} \\ 

\noalign{\smallskip}\hline\noalign{\smallskip}
\multicolumn{ 5}{c}{\bf{Northwestern-UCLA Multiview}} \\ 
\noalign{\smallskip}\hline\noalign{\smallskip}

AlexNet & fc7 & 4096 & 69.9 & 78.7 \\ 
ResNet-50 & pool5 & 2048 & 75.7 & 77.3 \\ 
GoogLeNet & pool5 & \bf{1024} & \bf{76.4} & \bf{79.8} \\ 

\noalign{\smallskip}\hline
\end{tabular}
\vspace{-2mm}
\end{table}


We determine the best CNN architecture that maximizes generalization power of HPM$_{\mathrm{RGB}}$ and HPM$_{\mathrm{3D}}$ for RGB and depth images respectively. We use the raw synthetic pose images to fine tune AlexNet, GoogLeNet, and ResNet-50, and then test them on the UWA and NUCLA datasets. Since the trained model is to be used as a frame-wise feature extractor for action recognition, we take recognition accuracy and feature dimensionality both into account. Table \ref{tab:models_compare} compares the average results on all possible cross-view combinations for the two datasets. The results show that for RGB videos, GoogLeNet and ResNet-50 perform much better than AlexNet. GoogLeNet also performs the best for depth videos and has the smallest feature dimensionality.  Therefore, we select GoogLeNet as the network architecture for both HPM$_{\mathrm{RGB}}$ and HPM$_{\mathrm{3D}}$ in the remaining experiments.

\subsection{Quantitative Analysis of GAN Refinement}

We quantitatively compare the effect of GAN-refinement using the UWA and NUCLA datasets by comparing the action recognition accuracies when the HPM$_{\mathrm{RGB}}$ and HPM$_{\mathrm{3D}}$ are fine tuned once on raw synthetic images and once on GAN-refined synthetic images. 

Table \ref{tab:GAN_compare} shows the average accuracies for all cross-view combinations on the respective datasets. We can see that the HPM$_{\mathrm{RGB}}$ fine tuned on the GAN-refined synthetic RGB images achieves 3.3\% and 1.4\% improvement over the one fine tuned with raw synthetic images on the UWA and NUCLA datasets respectively. For HPM$_{\mathrm{3D}}$, GAN-refined synthetic data also improves the recognition accuracy for the two datasets by 1\% and 1.3\% respectively. The improvements are achieved because the distribution gap between synthetic and real images is narrowed by GAN refinement.

\begin{table}[t]
\centering
\caption{Comparison of action recognition accuracy (\%) for GoogLeNet based HPM$_{\mathrm{RGB}}$ and HPM$_{\mathrm{3D}}$ when trained on raw synthetic images and GAN-refined synthetic images}
\label{tab:GAN_compare}
\begin{tabular}{lcc}
\hline\noalign{\smallskip}
Training Data & \multicolumn{1}{l}{HPM$_{\mathrm{RGB}}$} & \multicolumn{1}{l}{HPM$_{\mathrm{3D}}$} \\ 

\noalign{\smallskip}\hline\noalign{\smallskip}
\multicolumn{ 3}{c}{\bf{UWA3D Multiview Activity-II}} \\ 
\noalign{\smallskip}\hline\noalign{\smallskip}

Raw synthetic images & 64.7 & 73.8 \\ 
GAN-refined synthetic images & \bf{68.0} & \bf{74.8} \\ 

\noalign{\smallskip}\hline\noalign{\smallskip}
\multicolumn{ 3}{c}{\bf{Northwestern-UCLA Multiview}} \\ 
\noalign{\smallskip}\hline\noalign{\smallskip}

Raw synthetic images & 76.4 & 78.4 \\ 
GAN-refined synthetic images & \bf{77.8} & \textbf{79.7} \\ 

\hline\noalign{\smallskip}
\end{tabular}
\vspace{-2mm}
\end{table}

Recall that the real images used as a benchmark distribution for GAN refinement are neither from UWA nor NUCLA dataset. We impose no dependence on the type of real images used for SimGAN learning, because it is an unsupervised process and no pose labels are required. In the remaining experiments, we use  HPM$_{\mathrm{RGB}}$ and HPM$_{\mathrm{3D}}$ fine tuned with GAN refined synthetic images for comparison with other techniques.


\subsection{Comparison on the UWA3D Multiview-II Dataset}
\label{sec:UWA3D}

\begin{table*}[t]
\centering
\caption{Action recognition accuracy (\%) on the UWA3D Multiview-II dataset. $V_{1,2}^3$ means that view 1 and 2 were used for training and view 3 alone was used for testing. References for the existing methods are DVV\citep{DVV}, Action Tube\citep{Action_Tube}, CVP\citep{CVP}, LRCN\citep{LRCN}, AOG\citep{AOG}, Hankelets\citep{Hankelets}, JOULE\citep{hu2015jointly}, Two-stream\citep{simonyan2014two}, DT\citep{wang2011action}, C3D\citep{tran2015learning}, nCTE\citep{nCTE}, NKTM\citep{NKTM}, R-NKTM\citep{rahmani2017learning}. The symbol $\dagger$ indicates that the original model was fine-tuned with our synthetic data before applying the testing protocol.}
\setlength{\tabcolsep}{4.75pt}
\begin{tabular}{llccccccccccccc}
\hline\noalign{\smallskip}
Method & Data & $V_{1,2}^3$ & $V_{1,2}^4$ & $V_{1,3}^2$ & $V_{1,3}^4$ & $V_{1,4}^2$ & $V_{1,4}^3$ & $V_{2,3}^1$ & $V_{2,3}^4$ & $V_{2,4}^1$ & $V_{2,4}^3$ & $V_{3,4}^1$ & $V_{3,4}^2$ & Mean \\
\noalign{\smallskip}\hline\noalign{\smallskip}
\multicolumn{ 15}{c}{\textbf{Baseline}} \\ 
\noalign{\smallskip}\hline\noalign{\smallskip}
DVV & Depth & 35.4 & 33.1 & 30.3 & 40.0 & 31.7 & 30.9 & 30.0 & 36.2 & 31.1 & 32.5 & 40.6 & 32.0 & 33.7 \\ 

Action Tube & RGB & 49.1 & 18.2 & 39.6 & 17.8 & 35.1 & 39.0 & 52.0 & 15.2 & 47.2 & 44.6 & 49.1 & 36.9 & 37.0 \\ 

CVP & Depth & 36.0 & 34.7 & 35.0 & 43.5 & 33.9 & 35.2 & 40.4 & 36.3 & 36.3 & 38.0 & 40.6 & 37.7 & 37.3 \\

LRCN & RGB & 53.9 & 20.6 & 43.6 & 18.6 & 37.2 & 43.6 & 56.0 & 20.0 & 50.5 & 44.8 & 53.3 & 41.6 & 40.3 \\

AOG & RGB & 47.3 & 39.7 & 43.0 & 30.5 & 35.0 & 42.2 & 50.7 & 28.6 & 51.0 & 43.2 & 51.6 & 44.2 & 42.3 \\ 

LRCN$\dagger$ & RGB & 55.2 & 31.5 & 50.0 & 30.7 & 33.5 & 39.2 & 52.8 & 31.5 & 55.4 & 47.8 & 56.1 & 42.5 & 43.8 \\

Hankelets & RGB & 46.0 & 51.5 & 50.2 & 59.8 & 41.9 & 48.1 & 66.6 & 51.3 & 61.3 & 38.4 & 57.8 & 48.9 & 51.8 \\ 

JOULE & RGB-D & 43.6 & 67.1 & 53.6 & 64.4 & 56.4 & 49.1 & 65.7 & 48.2 & 76.2 & 33.5 & 79.8 & 46.4 & 57.0 \\ 
       
Two-stream & RGB & 63.0 & 47.1 & 55.8 & 60.6 & 53.4 & 54.2 & 66.0 & 50.9 & 65.3 & 55.5 & 68.0 & 51.9 & 57.6 \\

DT & RGB & 57.1 & 59.9 & 54.1 & 60.6 & 61.2 & 60.8 & 71.0 & 59.5 & 68.4 & 51.1 & 69.5 & 51.5 & 60.4 \\ 
     
C3D & RGB & 59.5 & 59.6 & 56.6 & 64.0 & 59.5 & 60.8 & 71.7 & 60.0 & 69.5 & 53.5 & 67.1 & 50.4 & 61.0 \\

nCTE & RGB & 55.6 & 60.6 & 56.7 & 62.5 & 61.9 & 60.4 & 69.9 & 56.1 & 70.3 & 54.9 & 71.7 & 54.1 & 61.2 \\ 

C3D$\dagger$ & RGB & 62.7 & 57.3 & 59.2 & 68.0 & 63.2 & 64.6 & 71.0 & 54.7 & 68.8 & 52.6 & 74.3 & 62.8 & 63.3 \\

NKTM & RGB & 60.1 & 61.3 & 57.1 & 65.1 & 61.6 & 66.8 & 70.6 & 59.5 & 73.2 & 59.3 & 72.5 & 54.5 & 63.5 \\ 

R-NKTM & RGB & 64.9 & 67.7 & 61.2 & 68.4 & 64.9 & 70.1 & 73.6 & 66.5 & 73.6 & 60.8 & 75.5 & 61.2 & \textbf{67.4} \\

\noalign{\smallskip}\hline\noalign{\smallskip}
\multicolumn{ 15}{c}{\textbf{Proposed}} \\
\noalign{\smallskip}\hline\noalign{\smallskip}

HPM$_{\mathrm{RGB}}$ & RGB & 72.4 & 73.4 & 64.3 & 71.9 & 50.8 & 62.3 & 69.9 & 61.8 & 75.5 & 69.4 & 78.4 & 66.2 & 68.0 \\ 

HPM$_{\mathrm{RGB}}$+Traj & RGB & 81.0 & 78.3 & 72.9 & 76.8 & 67.7 & 75.7 & 79.9 & 67.0 & 85.1 & 77.2 & 85.5 & 69.9 & 76.4 \\ 

HPM$_{\mathrm{3D}}$ & Depth & 80.2 & 80.1 & 75.6 & 78.7 & 59.0 & 69.0 & 72.1 & 65.2 & 84.8 & 79.1 & 82.5 & 71.1 & 74.8 \\ 

HPM$_{\mathrm{RGB}}$+HPM$_{\mathrm{3D}}$ & RGB-D & 79.9 & 83.9 & 76.3 & 84.6 & 61.3 & 71.3 & 77.0 & 68.9 & 85.1 & 78.7 & 87.0 & 74.8 & 77.4 \\

HPM$_{\mathrm{RGB}}$+HPM$_{\mathrm{3D}}$+Traj & RGB-D & 85.8 & 89.9 & 79.3 & 85.4 & 74.4 & 78.0 & 83.3 & 73.0 & 91.1 & 82.1 & 90.3 & 80.5 & \textbf{82.8}\\

\hline\noalign{\smallskip}
\end{tabular}
\label{tab:uwa_comp}
\end{table*}

Table \ref{tab:uwa_comp} compares our method with existing state-of-the-art. The proposed HPM$_{\mathrm{RGB}}$ alone achieves 68.0\% average recognition accuracy for RGB videos, which is higher than the nearest RGB-only competitor R-NKTM \citep{rahmani2017learning}, and for 8 out the 12 train-test combinations, our proposed HPM$_{\mathrm{RGB}}$ features provide significant improvement in accuracy. This shows that the invariant features learned by the proposed HPM$_{\mathrm{RGB}}$ are effective.

Combining HPM$_{\mathrm{RGB}}$ and dense trajectory features (Traj) gives a significant improvement in accuracy. It improves the RGB recognition accuracy to 76.4\%, which is 9\% higher than the nearest RGB competitor. It is also higher than the depth only method HPM$_{\mathrm{3D}}$. This shows that our method exploits the complementary information between the two modes of spatio-temporal representation, and enhances the recognition accuracy especially when there are large viewpoint variations. State-of-the-art RGB-D action recognition accuracy of 82.8\% on UWA dataset is achieved when we combine HPM$_{\mathrm{RGB}}$, HPM$_{\mathrm{3D}}$ and dense trajectory features.

In Table~\ref{tab:uwa_comp},  the results reported for the existing methods are achieved by using the original public models  and fine tuning them on UWA3D dataset under the used protocol. In contrast,  HPMs are not fine tuned to any dataset once  their training on the proposed synthetic data is completed. These  models are used out-of-the-box for the test data. These settings hold for all the experiments conducted in this work.  Indeed, fine tuning HPMs  on the real test datasets further improves the results but we avoid this step to show their generalization power.

Our data generation technique endows HPMs with inherent robustness to viewpoint variations along robustness to changes in  background, texture and clothing etc. The  baseline methods lack in these aspects which is a one of the reasons for the improvement achieved by our approach over those methods.  Note that, HPMs are unique in the sense that they model individual human poses in frames instead of actions. Therefore, our synthetic data generation method, which is an essential part of HPM training,  also focuses on generating human pose frames. One interesting enhancement of our data generation technique is to produce  synthetic videos instead. We can then  analyze the performance gain of (video-based) baseline methods  trained with our synthetic data. To explore this direction, we extended our technique to generate  synthetic videos and applied it to  C3D~\citep{tran2015learning} and LRCN~\citep{LRCN} methods as follows.

\begin{table*}
\floatbox[{\capbeside\thisfloatsetup{capbesideposition={left,top},capbesidewidth=5cm}}]{table}[\FBwidth]
{\caption{Action recognition accuracy (\%) on the NUCLA Multiview dataset. $V_{1,2}^3$ means that view 1 and 2 were used for training and view 3 was used for testing. The symbol $\dagger$ indicates that the original model was fine-tuned with our synthetic data before applying the testing protocol.}
\label{tab:nucla_comp}}
{
\begin{tabular}{llcccc}
\hline\noalign{\smallskip}
\multicolumn{ 1}{l}{Method} & \multicolumn{ 1}{l}{Data} & $V_{1,2}^3$ & $V_{1,3}^2$ & $V_{2,3}^1$ & \multicolumn{ 1}{c}{Mean} \\ 
\noalign{\smallskip}\hline\noalign{\smallskip}
\multicolumn{ 6}{c}{\textbf{Baseline}} \\
\noalign{\smallskip}\hline\noalign{\smallskip}

Hankelets~\citep{Hankelets} & RGB & 45.2 & - & - & 45.2 \\ 
JOULE~\citep{hu2015jointly} & RGB-D & 70.0 & 44.7 & 33.3 & 49.3 \\ 
LRCN~\citep{LRCN} & RGB & 64.0 & 36.2 & 51.7 & 50.6 \\
DVV~\citep{DVV} & Depth & 58.5 & 55.2 & 39.3 & 51.0 \\ 
LRCN$\dagger$ & RGB & 62.6 & 39.6 & 53.3 & 51.8 \\
CVP~\citep{CVP} & Depth & 60.6 & 55.8 & 39.5 & 52.0 \\ 
C3D~\citep{tran2015learning} & RGB & 71.2 & 53.7 & 54.5 & 59.8 \\ 
AOG~\citep{AOG} & Depth & 73.3 & - & - & - \\ 
C3D$\dagger$ & RGB & 68.4 & 64.6 & 53.2 & 62.1 \\
nCTE~\citep{nCTE} & RGB & 68.6 & 68.3 & 52.1 & 63.0 \\ 
NKTM~\citep{NKTM} & RGB & 75.8 & 73.3 & 59.1 & \textbf{69.4} \\ 
R-NKTM~\citep{rahmani2017learning} & RGB & 78.1 & - & - & - \\ 

\noalign{\smallskip}\hline\noalign{\smallskip}
\multicolumn{ 6}{c}{\textbf{Proposed}} \\ 
\noalign{\smallskip}\hline\noalign{\smallskip}

HPM$_{\mathrm{RGB}}$ & RGB & 91.5 & 69.0 & 73.1 & 77.8 \\ 
HPM$_{\mathrm{RGB}}$+Traj & RGB & 89.3 & 75.2 & 71.0 & 78.5 \\ 
HPM$_{\mathrm{3D}}$ & Depth & 91.9 & 75.2 & 71.9 & 79.7 \\ 
HPM$_{\mathrm{RGB}}$+HPM$_{\mathrm{3D}}$ & RGB-D & \textbf{92.4} & \textbf{74.1} & 76.8 & 81.1 \\ 
HPM$_{\mathrm{RGB}}$+HPM$_{\mathrm{3D}}$+Traj & RGB-D & 91.7 & 73.0 & \textbf{79.0} & \textbf{81.3} \\ 
\hline\noalign{\smallskip}
\end{tabular}
}
\end{table*}


For transparency, we  selected all `atomic' action sequences from CMU MoCap. Each of these sequences presents a single action, which also serves as the label of the video clip. To generate  synthetic videos, the frames in the training clips were processed according to the procedure described in Section~\ref{sec:data} and \ref{sec:GAN} with the following major differences. (1) No clustering was performed to learn representative poses because it was not required. (2) The parameters (i.e.  camera viewpoints, clothing etc.) were kept the same within a single synthetic video but  different  random settings were  adopted for each  video.  The size of the generated synthetic data  was matched to our ``pose''  synthetic data.  We took the  original C3D model that is pre-trained on the large scale dataset Sports-1M~\citep{karpathy2014large} and the original LRCN model that is pre-trained on UCF-101~\citep{soomro2012ucf101} and fine-tuned these models using our synthetic videos.  The fine-tuned models were then employed under the used protocol.

We report the results of these experiments in Table~\ref{tab:uwa_comp} by denoting our enhancements of C3D and LRCN as C3D$\dagger$ and LRCN$\dagger$. The results demonstrate that our synthetic data can improve the performance of baseline models for multi-view action recognition. The results also ascertain that  the proposed approach exploits the proposed data very effectively to achieve significant performance improvement over the existing methods.  We provide further discussion on the role of synthetic data in the overall performance of our approach in Section~\ref{sec:disc}.

\subsection{Comparison on the Northwestern-UCLA Dataset}
\label{sec:NUCLA}

Table \ref{tab:nucla_comp} comparative results on the NUCLA dataset. The proposed HPM$_{\mathrm{RGB}}$ alone achieves 77.8\% average accuracy which is 8.4\% higher than the nearest RGB competitor NKTM \citep{NKTM}. HPM$_{\mathrm{RGB}}$ +Traj further improves the average accuracy to 78.5\%. Our RGB-D method (HPM$_{\mathrm{RGB}}$+HPM$_{\mathrm{3D}}$+Traj) achieves 81.3\% accuracy which is the highest accuracy reported on this dataset.

\subsection{Comparison on the NTU RGB+D Dataset}

Table \ref{tab:ntu_rgbd_comp} compares our method with existing state-of-the-art on the NTU dataset. The proposed HPM$_{\mathrm{RGB}}$ uses RGB frames only and achieves 68.5\% cross-subject recognition accuracy, which is comparable to that of the best joints-based method ST-LSTM\citep{liu2016spatio} 69.2\% even though joints have been estimated from depth data and do not contain action irrelevant noises. This demonstrates that the HPM$_{\mathrm{RGB}}$ effectively learns features that are invariant to action irrelevant noises such as background, clothing texture and lighting etc.

\begin{table*}
\floatbox[{\capbeside\thisfloatsetup{capbesideposition={left,top},capbesidewidth=5cm}}]{table}[\FBwidth]
{\caption{Action recognition accuracy (\%) on the NTU RGB+D Human Activity Dataset. Our RGB only (HPM$_{\mathrm{RGB}}$+Traj) accuracies are higher than the nearest competitors which use RGB-D or Joints data. Our RGB-D method (HPM$_{\mathrm{RGB}}$+ HPM$_{\mathrm{3D}}$+Traj) outperforms all methods by significant margins in both settings}
\label{tab:ntu_rgbd_comp}}
{
\begin{tabular}{llcc}
\hline\noalign{\smallskip}
\multicolumn{1}{c}{} &  & Cross & Cross \\
Method & Data type & Subject & View \\
\noalign{\smallskip}\hline\noalign{\smallskip}
\multicolumn{ 3}{c}{\textbf{Baseline}} &  \\ 
\noalign{\smallskip}\hline\noalign{\smallskip}

HON4D~\citep{HON4D} & Depth & 30.6 & 7.3 \\ 
SNV~\citep{yang2014super} & Depth & 31.8 & 13.6 \\ 
HOG-2~\citep{ohn2013joint} & Depth & 32.4 & 22.3 \\ 
Skeletal Quads~\citep{evangelidis2014skeletal} & Joints & 38.6 & 41.4 \\ 
Lie Group~\citep{vemulapalli2014human} & Joints & 50.1 & 52.8 \\ 
Deep RNN~\citep{shahroudy2016ntu} & Joints & 56.3 & 64.1 \\ 
HBRNN-L~\citep{du2015hierarchical} & Joints & 59.1 & 64.0 \\ 
Dynamic Skeletons~\citep{hu2015jointly} & Joints & 60.2 & 65.2 \\ 
Deep LSTM~\citep{shahroudy2016ntu} & Joints & 60.7 & 67.3 \\ 
LieNet~\citep{huang2016deep} & Joints & 61.4 & 67.0 \\
P-LSTM~\citep{shahroudy2016ntu} & Joints & 62.9 & 70.3 \\ 
LTMD~\citep{luo2017unsupervised} & Depth & 66.2 & - \\
ST-LSTM~\citep{liu2016spatio} & Joints & 69.2 & \textbf{77.7} \\ 
DSSCA-SSLM~\citep{shahroudy2017deep} & RGB-D & \textbf{74.9} & - \\ 

\noalign{\smallskip}\hline\noalign{\smallskip}
\multicolumn{ 3}{c}{\textbf{Proposed}} & \multicolumn{1}{l}{} \\ 
\noalign{\smallskip}\hline\noalign{\smallskip}

HPM$_{\mathrm{RGB}}$ & RGB & 68.5 & 72.9 \\ 
HPM$_{\mathrm{RGB}}$+Traj & RGB & 75.8 & 83.2 \\ 
HPM$_{\mathrm{3D}}$ & Depth & 71.5 & 70.5 \\ 
HPM$_{\mathrm{RGB}}$+HPM$_{\mathrm{3D}}$ & RGB-D & 75.8 & 78.1 \\ 
HPM$_{\mathrm{RGB}}$+HPM$_{\mathrm{3D}}$+Traj & RGB-D & \bf{80.9} & \bf{86.1} \\ 

\hline\noalign{\smallskip}
\end{tabular}
}
\end{table*}

\vspace{2mm}
\noindent{\bf Comparison of RGB Results:}\\
Note that this paper is the first to provide RGB only human action recognition results on the challenging NTU dataset (see Table \ref{tab:ntu_rgbd_comp}). Our method (HPM$_{\mathrm{RGB}}$+Traj) outperforms all others by a significant margin while using only RGB data in both cross-subject and cross-view settings. In the cross-subject setting, our method achieves 75.8\% accuracy which is higher than state-of-the-art DSSCA-SSLM~\citep{shahroudy2017deep} even though DSSCA-SSLM uses both RGB and depth data whereas our method HPM$_{\mathrm{RGB}}$+Traj uses only RGB data. DSSCA-SSLM does not report cross-view results as it did not perform well in that setting~\citep{shahroudy2017deep} whereas our method achieves 83.2\% accuracy for the cross-view case which is 7.7\% higher than the nearest competitor ST-LSTM~\citep{liu2016spatio} which uses Joints data that is estimated from depth images. In summary, our 2D action recognition method outperforms existing 3D action recognition methods.

\vspace{2mm}
\noindent{\bf Comparison of RGB-D Results:}\\
From Table \ref{tab:ntu_rgbd_comp}, we can see that our RGB-D method (HPM$_{\mathrm{RGB}}$+ HPM$_{\mathrm{3D}}$+Traj) achieves state-of-the-art results in both cross-subject and cross-view settings outperforming the nearest competitors by 6\% and 8.4\% respectively.




\subsection{Timing}

\begin{table*}
\begin{center}
\caption{Execution time in seconds for the proposed method}
\label{tab:timing}
\begin{tabular}{llllllc}
\hline\noalign{\smallskip}
Data & \multicolumn{1}{l}{HPM Feature} & Trajectory & \multicolumn{1}{l}{FTP} & \multicolumn{1}{l}{SVM} & \multicolumn{1}{l}{Total} & \multicolumn{1}{l}{Rate(fps)} \\
\noalign{\smallskip}\hline\noalign{\smallskip}
RGB & 2.13E-02 & \multicolumn{1}{l}{2.78E-02} & 4.39E-05 & 2.95E-04 & 4.94E-02 & 20.2 \\
Depth & 2.13E-02 & - & 4.64E-05 & 3.32E-04 & 2.17E-02 & 46.1 \\
RGB-D & 2.13E-02 & \multicolumn{1}{l}{2.78E-02} & 4.64E-05 & 6.80E-04 & 4.98E-02 & 20.1 \\
\hline\noalign{\smallskip}
\end{tabular}
\end{center}
\end{table*}

Table \ref{tab:timing} shows the computation time for the major steps of our proposed method. Using a single core of a 3.4GHz CPU and the Tesla K-40 GPU, the proposed RGB \\ (HPM$_{\mathrm{RGB}}$+Traj) and RGB-D (HPM$_{\mathrm{RGB}}$+HPM$_{\mathrm{3D}}$+Traj) methods run at about 20 frames per second whereas the depth only method runs at about 46 frames per second. 

\section{Discussion}
\label{sec:disc}

When uncropped video frames are used to learn a neural network model, the background context is more dominant as it occupies more pixels. A recent study showed that by masking the human in the UCF-101 dataset, a 47.4\% ``human'' action recognition accuracy could still be achieved which, using the same algorithm, is only 9.5\% lower than when the humans are included \citep{he2016human}. Our HPM$_{\mathrm{RGB}}$ learns human poses rather than the background context which is important for surveillance applications where the background is generally static and any action can be performed in the same background. Moreover, HPM$_{\mathrm{RGB}}$ and HPM$_{\mathrm{3D}}$ are not fine tuned on any of the datasets on which they are tested. Yet, our models outperform all existing methods by a significant margin. For applications such as robotics and video retrieval where the background context is important, our HPM models can be used to augment the background context. For optimal performance, the cropped human images must be passed through the HPM$_{\mathrm{RGB}}$ and HPM$_{\mathrm{3D}}$. However, both HPMs are robust to cropping errors as many frames in the UWA dataset (especially view 4 in Fig.~\ref{fig:uwa3d_sample}) and the NTU dataset have cropping errors.

\vspace{-3mm}

\subsection{Comparison with existing synthetic data}
One of the major contributions of this work is synthetic data generation for robust action recognition. We note that SURREAL (Synthetic hUmans foR REAL tasks) \citep{varol17} is another recent method to generate synthetic action data that can be used to train the proposed HPMs. However, there are some major differences between SURREAL and the proposed synthetic data. (1) SURREAL was originally  proposed for body segmentation and depth estimation whereas our  dataset aims at modeling distinctive human poses from multiple viewpoints. While both datasets provide sufficient variations in clothing, human models, backgrounds, and illuminations; our dataset systematically covers 180$^{\rm o}$ of view to enable  viewpoint invariance, which is not the case for SURREAL. (2) To achieve realistic viewpoint variations, our dataset uses 360$^{\rm o}$ spherical High Dynamic Range Images whereas SURREAL uses the LSUN dataset~\citep{journals/corr/YuZSSX15} for backgrounds. Hence, our approach is more suitable for  large viewpoint variations. (3) Finally, we use Generative Adversarial Network to reduce the distribution gap between synthetic and real data. Our results in Table~\ref{tab:GAN_compare} already verified that this provides additional boost to the action recognition performance.

To demonstrate the use of SURREAL with our pipe-line and  quantitatively  analyze the advantages of the proposed dataset for robust action recognition, we compare the performance of our underlying approach using the two datasets on UWA3D and NUCLA databases.  
We repeated our experiments  using SURREAL as follows. First, we computed 339 representative poses from SURREAL using the HDBSCAN algorithm \citep{mcinnes2017hdbscan}  and  used the skeletal distance function in Eq.~(\ref{eq:1}) to assign frames in the dataset to these poses. 
HPMs are then trained on these frames using the 339 pose labels, followed by temporal encoding and classification. This pipeline is exactly the same as the one used for our data in Section~\ref{sec:Exp}, except that the representative poses are now  computed using SURREAL.

\begin{table}[t]
\centering
\caption{Action recognition accuracy (\%) for HPM$_{\mathrm{RGB}}$ and HPM$_{\mathrm{3D}}$ when trained with SURREAL dataset~\citep{varol17} and the proposed synthetic data. $V^{\text{challenge}}$ represents the most challenging viewpoint in the multi-view test data, i.e.  View 4 for UWA3D and View 3 for NUCLA dataset.}
\label{tab:SURREAL_compare}
\begin{tabular}{llcc}
\hline\noalign{\smallskip}
Method & Training Data & \multicolumn{1}{l}{$V^{\text{challenge}}$} & \multicolumn{1}{l}{Mean}\\ 

\noalign{\smallskip}\hline\noalign{\smallskip}
\multicolumn{ 4}{c}{\bf{UWA3D Multiview Activity-II}} \\ 
\noalign{\smallskip}\hline\noalign{\smallskip}

HPM$_{\mathrm{RGB}}$ & SURREAL & 61.6 & 67.4\\ 
HPM$_{\mathrm{RGB}}$ & Proposed data & {\bf 69.0} & \bf{68.0} \\ 
HPM$_{\mathrm{3D}}$ & SURREAL & 65.8 & 72.1\\ 
HPM$_{\mathrm{3D}}$ & Proposed data & {\bf 74.7} & \bf{74.8} \\ 

\noalign{\smallskip}\hline\noalign{\smallskip}
\multicolumn{ 4}{c}{\bf{Northwestern-UCLA Multiview}} \\ 
\noalign{\smallskip}\hline\noalign{\smallskip}

HPM$_{\mathrm{RGB}}$ & SURREAL & 69.9 & 74.4 \\ 
HPM$_{\mathrm{RGB}}$ & Proposed data & {\bf 73.1} & \textbf{77.8} \\ 
HPM$_{\mathrm{3D}}$ & SURREAL & 68.1 & 77.4 \\ 
HPM$_{\mathrm{3D}}$ & Proposed data & {\bf 71.9} & \textbf{79.7} \\ 

\hline\noalign{\smallskip}
\end{tabular}
\end{table}

Table~\ref{tab:SURREAL_compare} reports the mean recognition accuracies of HPMs on UWA3D and NUCLA  datasets  when  trained using the SURREAL dataset and the proposed data. 
The table also includes  results for the most challenging viewpoints in the datasets. For UWA3D, View~4  is challenging   due to the large variations in both azimuth and elevation angles (see Fig.~\ref{fig:uwa3d_sample}). For NUCLA, View~3  is particularly challenging as compared to the other viewpoints (see Fig.~\ref{fig:nucla_sample}).   From the results, we can see that the proposed data is able to achieve significant performance gain over SURREAL for these viewpoints. 
In our opinion,  systematic coverage of  180$^{\rm o}$ of view in our data is the main reason behind this fact.
Our data also achieves a consistent overall gain for both RGB and depth modalities of the test datasets.

\subsection{Improvements with synthetic data}
Our experiments in Section~\ref{sec:Exp} demonstrate a significant performance gain over the current state-of-the-art. Whereas the contribution of the network architecture, data modalities and GAN to the overall performance is clear from the presented experiments, we further investigate the performance gain contributed by the proposed synthetic data. To that end, we  first compare HPM$_{\mathrm{RGB}}$, which has been fine-tuned with the proposed data, to HPM$_{\mathrm{Orig}}$ which is the original GoogLeNet model  - not fine-tuned with our synthetic data. To ensure a fair comparison, all the remaining steps in the proposed pipeline, including  temporal encoding and classification, are kept exactly the same for the two cases.  The first two rows of Table~\ref{tab:improvement_compare} compare the mean recognition accuracies of  HPM$_{\mathrm{RGB}}$ and HPM$_{\mathrm{Orig}}$ for  UWA3D and NUCLA datasets. 
These results ascertain a clear performance gain with the proposed synthetic dataset.

The last two rows of Table~\ref{tab:improvement_compare} examine the performance gain of two popular  baseline methods when fine-tuned on our synthetic data. Although significant, the average  improvement in the accuracies of these methods is rather small compared to that of our method on our synthetic data (first two rows). Recall that our data generation method generates synthetic ``poses'' rather than videos and we had to extend our method to generate synthetic videos for the sake of this experiment. 
Details on synthetic video generation and training of the baseline methods are already provided in Section~\ref{sec:UWA3D}. 
From the results in Table~\ref{tab:improvement_compare}, we conclude that our proposed method exploits our synthetic data more effectively, and both the proposed method and our synthetic data contribute significantly to the overall performance gain.

\begin{table}[t]
\centering
\caption{Analysis of performance gain due to the proposed data and method. HPM$_{\text{Orig}}$ is the original GoogLeNet. The symbols $\dagger$ and $\uparrow$ denote  enhancement of the existing techniques with our  data and the improvement in accuracy (\%), respectively.}
\label{tab:improvement_compare}
\begin{tabular}{llcc}
\hline\noalign{\smallskip}
Method & Training & \multicolumn{1}{l}{UWA3D} & \multicolumn{1}{l}{NUCLA} \\ 

\noalign{\smallskip}\hline\noalign{\smallskip}

HPM$_{\mathrm{Orig}}$ & without synthetic data & 62.8 & 66.7 \\ 
HPM$_{\mathrm{RGB}}$ & with synthetic data & \bf{68.0} & \textbf{77.8} \\ 

\noalign{\smallskip}\hline\noalign{\smallskip}


C3D$\dagger$ & with synthetic data &$\uparrow$2.3 & $\uparrow$2.3\\ 
LRCN$\dagger$ & with synthetic data & $\uparrow$3.5 & $\uparrow$1.2\\

\hline\noalign{\smallskip}
\end{tabular}
\end{table}

\subsection{Role of synthetic RGB data in action recognition}
Although our approach deals with RGB, depth and RGB-D data;  we find it necessary to briefly discuss the broader role of synthetic RGB data in human action recognition. In contrast to depth videos, multiple large scale RGB video action datasets are available to train deep action models. Arguably, this diminishes the need of synthetic data in this domain.
However, synthetic data generation methods such as ours and \citep{varol17} are able to easily ensure a wide variety of action irrelevant variations in the data, e.g. in camera viewpoints, textures, illuminations; up to any desired scale.  In natural videos, such variety and scale of variations can not be easily guaranteed even in large scale datasets.  For instance,  in our experiments in Sections~\ref{sec:UWA3D} and \ref{sec:NUCLA}, both C3D and LRCN were originally pre-trained on large scale RGB video datasets in Table~\ref{tab:uwa_comp} and \ref{tab:nucla_comp}, yet our RGB synthetic data was able to boost their  performance.  Our synthetic data method easily and efficiently captures as many variations of  the exact same action as desired, a real-world analogous to which is extremely difficult.

We also tested the  performance of our approach on UCF-101 dataset~\citep{soomro2012ucf101} to analyze the potential of synthetic data and HPMs for the standard action recognition benchmarks in the RGB domain. UCF-101 is a popular RGB-only action dataset, which includes video clips of 101 action classes. The actions covers 1) Human-Object Interaction, 2) Human Body Motion, 3) Human-Human Interaction, 4) Playing Musical Instruments, and 5) Sport Actions. Since we trained our HPMs to model human \emph{poses}, the appearances of human poses in the test videos are important for a transparent analysis.  Therefore, we selected 1910 videos from the dataset with 16 classes of Human Body Motion, and classified them  using the proposed HPMs.
Table~\ref{tab:ucf101_compare} reports the performance of our approach along the accuracy  of C3D on the same subset for comparison. The table also reports the accuracy of HPM$_{\text{RGB}}\dagger$, for which we used twice as many pose labels and training images as used for training HPM$_{\text{RGB}}$.  
 This improved the performance of  our approach, indicating the advantage of easily producible synthetic data. Notice that, whereas  the performance of HPMs remains comparable to C3D, the latter is trained on Millions of `videos' as compared to the few hundred thousand `frames' used for training our model.  Moreover, our model is nearly 7 times smaller than C3D   in size.  These facts clearly demonstrate the usefulness of the proposed method and synthetic data generation technique for standard RGB action recognition.

\begin{table}[t]
\centering
\caption{Action recognition accuracy (\%) on the Human Body Motion subset of  UCF-101 dataset. For a transparent analysis, the results do not include augmentation by trajectory features and/or ensemble features for any of the approaches. The symbol $\dagger$ denotes larger (2$\times$) synthetic training data size. }
\label{tab:ucf101_compare}
\begin{tabular}{lc}
\hline\noalign{\smallskip}
Method & \multicolumn{1}{l}{Human Body Motion} \\ 
\noalign{\smallskip}\hline\noalign{\smallskip}

C3D & 84.8 \\ 
HPM$_{\mathrm{RGB}}$ & 82.5 \\ 
HPM$_{\mathrm{RGB}}\dagger$ & 84.6 \\ 

\hline\noalign{\smallskip}
\end{tabular}
\end{table}

\section{Conclusion}
We proposed Human Pose Models for human action recognition in RGB, depth and RGB-D videos. The proposed models uniquely represent human poses irrespective of the camera viewpoint, clothing textures, background and lighting conditions. We proposed a method for synthesizing realistic RGB and depth training data for learning such models. The proposed method learns 339 representative human poses from MoCap skeleton data and then fits 3D human models to these skeletons. The human models are then rendered as RGB and depth images from 180 camera viewpoints where other variations such as body shapes, clothing textures, backgrounds and lighting conditions are applied. We adopted Generative Adversarial Networks (GAN) to reduce the distribution gap between the synthetic and real images. Thus, we were able to generate millions of realistic human pose images with known labels to train the Human Pose Models. The trained models contain complementary information between RGB and depth modalities, and also show good compatibility to the hand-crafted dense trajectory features. Experiments on three benchmark RGB-D datasets show that our method outperforms existing state-of-the-art on the challenging problem of cross-view and cross-person human action recognition by significant margins. The HPM$_{\mathrm{RGB}}$, HPM$_{\mathrm{3D}}$ and Python script for generating the synthetic data will be made public.


%



\begin{acknowledgements}
This research was sponsored by the Australian Research Council grant DP160101458. The Tesla K-40 GPU used for this research was donated by the NVIDIA Corporation.
\end{acknowledgements}

\balance

\bibliographystyle{myspbasic}      
\bibliography{jianl_journal}   

\end{document}